# A Multi-directional Meta-Learning Framework for Class-Generalizable Anomaly Detection

**Padmaksha Roy**[1] , **Lamine Mili**[2] , **Almuatazbellah Boker**[2,3]

[1] [2] [3] Virginia Tech

{padmaksha,lmili,boker}@vt.edu

## Abstract

In this paper, we address the problem of class-generalizable anomaly detection, where the objective is to develop a unified model by focusing our learning on the available normal data and a small amount of anomaly data in order to detect the completely unseen anomalies, also referred to as the out-of-distribution (OOD) classes. Adding to this challenge is the fact that the anomaly data is rare and costly to label. To achieve this, we propose a multidirectional meta-learning algorithm – at the inner level, the model aims to learn the manifold of the normal data (representation); at the outer level, the model is meta-tuned with a few anomaly samples to maximize the softmax confidence margin between the normal and anomaly samples (decision surface calibration), treating normals as in-distribution (ID) and anomalies as out-of-distribution (OOD). By iteratively repeating this process over multiple episodes of predominantly normal and a small number of anomaly samples, we realize a multidirectional meta-learning framework. This two-level optimization, enhanced by multidirectional training, enables stronger generalization to unseen anomaly classes.

## 1 Introduction

Anomaly detection (AD) has a wide spectrum of applications in various domains. In industrial settings, it is used to monitor machinery and detect early signs of equipment failure or faults. In network traffic and cybersecurity, anomaly detection identifies malicious activity, intrusions, or deviations in system logs that may indicate attacks or breaches. In climate science, it helps uncover unusual patterns in weather data, environmental monitoring, or long-term climate trends. In healthcare, anomaly detection supports early diagnosis by flagging irregularities in patient records, physiological signals, or medical imaging.

In anomaly detection, anomalies are inherently rare, heterogeneous, and constantly evolving, which makes them difficult to characterize exhaustively during training without additional retraining or fine-tuning. By contrast, normal data are abundant and comparatively stable across datasets, making them a more reliable foundation for representation learning. Putting emphasis on learning *the normal data manifold* allows capturing consistent structures and regularities that define *"normality"* across domains. Once this manifold is well learned, even small deviations caused by anomalies become more distinguishable, enabling the detection of novel or unseen attacks. Furthermore, fine-tuning with a limited set of anomalies can then act as a corrective step, improving sensitivity without requiring exhaustive coverage of all possible anomaly types.

This makes class-generalizable anomaly detection a particularly challenging and important problem to address. Adding to this challenge is the fact that normal classes vary significantly between datasets. In some cases, the feature correlation of anomaly datasets lies very close to the normal data, making them very difficult to separate. As a result, normal data from previously unseen domains can often be misclassified as anomalies, leading to high false positive rates. Our goal in this paper is to design a class-generalizable few-shot and zero-shot anomaly detection algorithm that learns robust latent features from multiple classes and can generalize to entirely unseen or out-of-distribution (OOD) domains. ODIN [Liang *et al.*, 2017] observes that in-distribution (ID) samples typically exhibit a larger softmax confidence score gradient than most OOD samples.

The idea of meta-learning involves training in episodes that simulate deployment: in each episode, we (a) meta-train on some source domains and (b) meta-test on a held-out domain from the same pool, updating the model so it does well on both. Repeating this biases the representation toward domain-invariant cues that transfer to unseen domains at test time. Because every update is judged by performance on a different domain within the episode, the network learns features that survive domain swaps (i.e., are stable across shifts) and de-emphasizes spurious, domain-specific correlations. MAML [Finn *et al.*, 2017; Hospedales *et al.*, 2021] learns a model initialization that can adapt to a new task with just a few gradient steps. It optimizes a meta-objective across many tasks so that the post-adaptation loss is low. For OOD domain generalization, we form tasks from different domains and meta-train so that the model adapts quickly to a held-out domain.

Multi-direction meta-learning is a concept within the field of meta-learning that involves transferring knowledge in multiple directions to improve a model's performance on new tasks. This approach is particularly effective for few-shot and zero-shot learning, where models must adapt to new tasks using only a small number of examples. Multi-direction meta-learning distinguishes itself by not just transferring knowledge in a single direction (e.g., from a pre-trained model to a new task). Instead, it enables a more collaborative and dynamic exchange of information. In contrast to a one-way knowledge transfer from a fixed source to a new task, multi-direction meta-learning allows different parts of the learning system to "teach" each other to achieve a common goal.

Basically, in our formulation, a *"direction"* corresponds to a particular ID(normal)–vs–OOD(anomaly) partition of families: an inner set of normal families $\mathcal{C}_{\text{ID}}^{(e)}$ and an outer set of anomaly families $\mathcal{C}_{\text{meta}}^{(e)}$ used in episode $e$. Each episode samples a different pairing $(\mathcal{C}_{\text{ID}}^{(e)}, \mathcal{C}_{\text{meta}}^{(e)})$, so the inner loss learns representations from one subset of normal domains, while the outer loss calibrates the decision boundary using a small subset of anomaly domains. Over many episodes we sweep through many such pairings, effectively training across multiple "directions" of transfer (e.g., normal$_A \rightarrow$ anomaly$_B$, normal$_C \rightarrow$ anomaly$_D$, etc.), rather than a single fixed source–target split. Although algorithmically, the method is still a bilevel meta-learning procedure over multiple tasks, the distinctive part lies in *how tasks are constructed*: inner episodes always aggregate multiple normal families (rich ID support), outer episodes use few-shot anomaly supports from different OOD families, and the families used in outer meta-training are disjoint from those used as held-out OOD at test time. This repeated re-pairing of different normal and anomaly subsets is what we intended by *"multi-directional"* flow of information and allows multiple interacting axes of adaptation.

Building on these two principles, we design a two-stage representation learning algorithm: first, we learn the normal manifold across multiple normal domains in a shared latent space; then, we meta-tune the representation using a small number of anomaly samples to separate in-distribution (normal) from out-of-distribution (anomalies) in their softmax confidence space. Moreover, the latent space of a classifier compresses raw inputs into task-relevant features, thereby removing spurious class-specific information [Muthukrishna and Gupta, 2024; Wang *et al.*, 2025; Angiulli *et al.*, 2023]. Well-structured latent spaces cluster normal patterns and push atypical ones into low-density regions. Therefore, calibrating and regularizing this latent space directly improves robustness to distribution shifts. Moreover, the presence of a complex latent space with multiple, heterogeneous anomaly clusters makes it significantly harder for a single detection model to perform well in a multi-class setting.

In summary, our contribution can be summarized as follows:

- We address the challenge of OOD generalization in the context of anomaly detection (AD) by proposing a bi-level meta-learning algorithm that explicitly *disentangles* representation learning from decision calibration. The inner level learns *the manifold of the normal data* across multiple domains, while the outer level *meta-tunes this boundary* with small numbers of anomaly samples to further distinguish normal from anomalies in their softmax confidence space.

- The framework is extended to *multiple meta directions* by constructing episodes that combine normal data (at inner level) with a minimal set of anomalies (at outer level) from different domains. This multitask objective relaxes the dependency on class-specific source and targets, allowing multiple independent, interacting axes of adaptation.

- In this study, our goal is to improve OOD anomaly detection(AD) for both the train anomaly classes (used for meta-tuning) and the completely unseen (test) anomalies. Experiments demonstrate consistent gains across key classification metrics– Precision, Recall, AUC, when meta-tuning with anomaly classes are used.

## 2 Related Work

Domain generalization methods are typically grouped into four main categories: domain-invariant representation learning, meta-learning, latent dimension regularization, and metric learning. The first group aims to extract features that remain stable across domains, enabling transfer to unseen settings - for example, autoencoder-based methods [Ghifary *et al.*, 2015] use multidomain training with augmentation to learn shared representations, MMD-AAE [Li *et al.*, 2018] aligns heterogeneous distributions through adversarial learning, while approaches like domain-specific masking [Chattopadhyay *et al.*, 2020], noise-enhanced autoencoders [Liang *et al.*, 2021] and moment-alignment techniques [Jin *et al.*, 2020; Lu *et al.*, 2022] capture both invariant and discriminative features. The second group, metalearning, improves generalization by leveraging related tasks, such as latent space projections to mitigate domain bias [Erfani *et al.*, 2016], MAML-style methods that adapt updates in latent space [Finn *et al.*, 2017; Rusu *et al.*, 2018], and zero-shot learning [Wang *et al.*, 2019] that transfers knowledge from seen to unseen classes. A third direction, based on information bottleneck and metric learning, emphasizes disentangling spurious correlations, with theoretical guarantees provided by the variance–invariance–covariance framework [Shwartz-Ziv *et*

*al.*, 2023], although adversarial adaptation methods remain sensitive to distribution mismatches. Early work like [Hendrycks and Gimpel, 2016] relied on softmax statistics for error and OOD detection, later extended by [Xu *et al.*, 2018] through unsupervised adaptation that minimizes inter-domain discrepancies, while more recent methods such as nearest-neighbor–based detection [Sun *et al.*, 2022] and causal invariant learning [Arjovsky *et al.*, 2019] provide flexible distribution-agnostic strategies for cross-domain generalization. The ODIN paper Liang *et al.* [Liang *et al.*, 2017] shows that the standard softmax confidence is unreliable for detecting OOD inputs because networks often assign high probabilities to unfamiliar data. It proposes two simple test-time tweaks—temperature scaling of the softmax and a small input perturbation that amplify the confidence gap between ID and OOD samples. With these, ID examples become more confident, while OOD ones become less confident, enabling effective OOD detection using a basic threshold on the scaled softmax score. MTL-RED [Roy and Choi, 2025] proposes a novel classification framework that leverages regularization techniques to guide the latent space toward retaining only the most relevant features for out-of-distribution (OOD) classification. The result is a compressed, invariant representation that effectively discards *spurious* domain-specific information. The paper [Yao *et al.*, 2025] addresses class-generalizable AD by proposing ResAD, a framework that detects anomalies in unseen classes without retraining. The key idea is to model residual features instead of raw features, which reduces interclass variation and keeps normal residuals consistent across classes. ResAD integrates a feature converter, feature constraintor, and feature distribution Estimator to learn stable residual distributions, making anomalies identifiable as out-of-distribution. MetaOOD [Qin *et al.*, 2024] is a zero-shot, unsupervised framework that uses metalearning to automatically select the most suitable OOD detection model for a new dataset without requiring labels. It leverages historical performance data and language-model embeddings of datasets and models to capture task similarities, achieving superior reliability across diverse domains. They propose an autoencoder-based meta-learning method that adapts anomaly detectors to new, unlabeled target tasks without assuming "almost all unlabeled data is normal," which often fails in practice. The key idea is to learn per-instance anomalous attributes inside the reconstruction loss (so anomalies aren't reconstructed) and to adapt only the AE's last layer + these attributes via an efficient closed-form alternating update, validated on four real datasets. The authors [Kumagai *et al.*, 2023] propose a meta-learned autoencoder adapts to new unlabeled tasks by learning per-sample "anomalous attributes" that stop anomalies from being reconstructed, using an efficient closed-form alternating update. MetaLog [Zhang *et al.*, 2024] uses a shared semantic embedding space plus meta-learning to transfer anomaly detection from mature systems to new ones.

## 3 Problem Formulation

Let $\mathcal{C} = \{1, 2, \ldots, K\}$ denote the set of class labels ("families"). We consider a collection of tasks $\mathcal{I} = \{I_1, \ldots, I_Q\}$, where each task $I_q \subseteq \mathcal{C}$ specifies a subset of labels used to define a domain split. For a designated reference task $I_1$, write its complement $I_1^c = \mathcal{C} \setminus I_1$. We are given a dataset $\mathcal{D} = \{(x, y)\}$ with $x \in \mathcal{X}$ and $y \in \mathcal{C}$, drawn from mixtures whose supports align with the sets in $\mathcal{I}$. We split $I_1^c$ into two disjoint subsets $I_{\text{meta}}$ and $I_{\text{held}}$. $I_{\text{meta}}$ contains anomaly families used as OOD during outer meta-training, while $I_{\text{held}}$ contains anomaly families used only at evaluation time and are never included in any inner or outer episode, $\mathcal{C}_{\text{meta-OOD}} \cap \mathcal{C}_{\text{held}} = \varnothing$.

Let $f_\theta : \mathcal{X} \to \mathcal{Z}$ be an encoder and $h_\phi : \mathcal{Z} \to \mathbb{R}^K$ a classifier producing logits $\ell(x) = h_\phi(f_\theta(x)) \in \mathbb{R}^K$ for classes $\{1, \ldots, K\}$. For a temperature $T > 0$, we define the softmax score for class $i$ as

$$S_i(x; T) = \frac{\exp([\ell(x)]_i/T)}{\sum_{j=1}^{K} \exp([\ell(x)]_j/T)}. \quad (1)$$

The predicted label is $\hat{y}(x) = \arg\max_i S_i(x; T)$, and the *maximum softmax probability* (confidence) is

$$S_{\hat{y}}(x; T) = \max_i S_i(x; T). \quad (2)$$

**Bilevel Meta-learning Objective** We aim to learn the encoder-decoder parameter ($\theta$) that learns the representation of the data, the softmax decision surface calibration parameter ($\phi$) and adjust the temperature scaling parameter ($T$) with the final goal being generalizing across domain splits and separating benign classes from both training OOD(used for meta-tuning) and the unseen test anomalies (test OOD) classes. The inner loop models only ID normals, while *OOD anomalies used for meta-tuning appear only in the outer loop*. Confidence calibration is handled by $\phi$ and to a minor extent by $T$ at the outer meta level.

*Inner (task) objective: Normals-only one-class binary cross-entropy (BCE).* Given an ID distribution $\mathcal{P}_{\text{in}}$ supported on $I_1^c$, we adapt the encoder by maximizing confidence in normal data via a one-class binary cross-entropy (all targets fixed to 1):

$$\theta^\star(\phi) = \arg\min_\theta \mathbb{E}_{x \sim \mathcal{S}_{\text{id}}} \Big[ \text{BCE}\big(S_y(x; T, \theta, \phi), 1\big) \Big]. \quad (3)$$

We experiment with both fixing and learning $T$ in the inner step as a simple case.

*Outer (meta) objective:* In this stage, we employ small number of anomaly(meta-OOD) samples to meta-tune the model to distinguish normal and anomaly samples. Let $\mathcal{P}_{\text{in}}$ denote the ID-normal distribution and $\mathcal{P}_{\text{out}}$ an anomaly(OOD) distribution. Holding $\theta = \theta^\star(\phi)$ fixed, we minimize a calibrated BCE objective over $(\phi, T)$ to widen the ID–OOD confidence gap or otherwise we encourage high confidence on ID (normals) and low confidence on OOD(anomalies). Figure 1 explains our framework briefly.

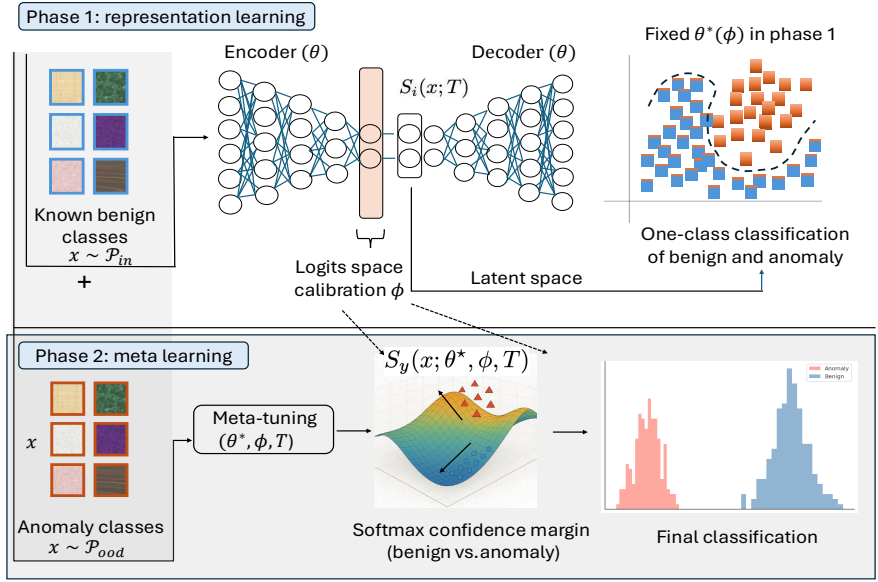

Figure 1: The Multi-directional Meta-learning Framework. In Phase 1, we train the model with a One-Class Classifier consisting of known normal classes in order to learn the manifold of the normal data representation. In Phase 2, we meta-tune the softmax decision boundary with a small number of anomaly(OOD) samples to push the softmax confidence margin between ID(normal) and OOD(anomaly).

$$\min_{\phi, T} \quad \mathcal{L}_{\text{meta}}(\phi, T; \theta^\star) = \mathcal{L}_{\text{id}}(\phi, \theta^\star, T)$$
$$+ \mathcal{L}_{\text{meta-ood}}(\phi, \theta^\star, T) \quad (4)$$
$$+ \alpha\, \mathcal{L}_{\text{margin}}(\phi, \theta^\star, T)$$
$$\text{s.t.} \quad \theta^\star(\phi) \text{ defined in Eq. (4)}.$$

The BCE terms are

$$\mathcal{L}_{\text{id}}(\phi, \theta^\star, T) = \mathbb{E}_{x \sim \mathcal{P}_{\text{id}}}\big[\text{BCE}\big(S_y(x; \theta^\star, \phi, T), 1\big)\big],$$
$$\mathcal{L}_{\text{meta-ood}}(\phi, \theta^\star, T) = \mathbb{E}_{x \sim \mathcal{P}_{\text{ood}}}\big[\text{BCE}\big(S_y(x; \theta^\star, \phi, T), 0\big)\big] \quad (5)$$

and the margin hinge is

$$\widehat{g} = \mathbb{E}_{x \sim \mathcal{P}_{\text{id}}}\big[S_y(x; \theta^\star, \phi, T)\big]$$
$$- \mathbb{E}_{x \sim \mathcal{P}_{\text{meta-ood}}}\big[S_y(x; \theta^\star, \phi, T)\big],$$
$$\mathcal{L}_{\text{margin}}(\phi, \theta^\star, T) = \big[m - \widehat{g}\big]_+. \quad (6)$$

where, $m > 0$ is a *fixed* margin hyperparameter (ID–OOD logit gap), $\alpha$ is the weightage put on the margin value. $S_y(\cdot)$ denotes the calibrated normal and anomaly class logit (pre-sigmoid) evaluated at the input $x$.

## 4 Experiment Settings

In this section, we demonstrate the performance of our proposed model on benchmark cybersecurity and healthcare datasets.

Equations (3)–(5) explicitly define a bi-level formulation that *separates the ID–OOD confidence margin* by minimizing a logistic loss on the softmax score $s(\cdot)$ w.r.t. $(\phi, T)$, while the inner solution $\theta^\star(\phi)$ (learned on normals only) fixes the normal representation manifold. This makes the intended margin widening precise and optimizable.

**Episodic Multi-task Extension** To emulate deployment across multiple domain splits, we form episodes over $I_q \in \mathcal{I}$, treating $I_q^c$ as ID and $I_q$ as OOD from different anomaly classes per episode. Let $\mathcal{P}_{\text{in}}^{(q)}$ and $\mathcal{P}_{\text{out}}^{(q)}$ be the corresponding episode distributions.

The overall multidirectional meta-objective aggregates episodes,

$$\min_{\phi, T} \frac{1}{Q} \sum_{q=1}^{Q} \mathcal{L}_{\text{meta}}^{(q)}(\phi, \theta, T). \quad (7)$$

The number of episodes is a training hyperparameter that controls how many times we resample and revisit domain splits. More episodes give a better Monte-Carlo approximation of the summed meta-objective. We maintain a single set of encoder–decoder–classifier parameters $\theta$ that is shared across episodes. Each episode performs several inner updates on ID (normal) families starting from the current $\theta$, and the resulting $\theta$ is carried over to the next episode. Similarly, for the outer meta-calibration loss, the classifier weights for $(\phi, T)$ are updated cumulatively across episodes.

### 4.1 Training Strategy

We train in episodes that mimic a train–test shift: each episode selects a set of ID (normal) families for the inner loop and a few-shot set of OOD families for the outer

Table 1: Performance metrics (Precision (P), Recall (R), AUC, F1) for various methods across attack datasets. Each attack dataset (balanced) is considered as held-out or meta-tune based on what stage they have been used. While held-out OODs are completely unseen, meta-tuning OODs consist of different combinations of the other attack datasets used in the outer loss (5–50 samples from each attack class). CSE-CIC-IDS dataset(first two rows BOTNET-INFILTRATION), combined CIC-IOT/IoMT datasets (last two rows DDOS-WEB).

| Method | BOTNET (meta-tune) | | | | SSH-BRUTEFORCE (meta-tune) | | | | GOLDENEYE (meta-tune) | | | |
|---|---|---|---|---|---|---|---|---|---|---|---|---|
| | P | R | AUC | F1 | P | R | AUC | F1 | P | R | AUC | F1 |
| CORAL | 0.706 ± 0.01 | 0.467 ± 0.01 | 0.698 ± 0.01 | 0.562 ± 0.01 | 0.598 ± 0.01 | 0.860 ± 0.01 | 0.785 ± 0.01 | 0.705 ± 0.02 | 0.545 ± 0.01 | 0.448 ± 0.01 | 0.551 ± 0.01 | 0.492 ± 0.01 |
| MTAE | 0.545 ± 0.01 | 0.448 ± 0.01 | 0.551 ± 0.01 | 0.492 ± 0.01 | 0.653 ± 0.01 | 0.860 ± 0.01 | 0.791 ± 0.01 | 0.797 ± 0.01 | 0.667 ± 0.01 | 0.386 ± 0.01 | 0.481 ± 0.01 | 0.581 ± 0.01 |
| MTL-RED | 0.653 ± 0.01 | 0.860 ± 0.01 | 0.791 ± 0.01 | 0.742 ± 0.01 | **0.930 ± 0.01** | 0.724 ± 0.01 | 0.872 ± 0.01 | 0.814 ± 0.01 | **0.916 ± 0.01** | 0.804 ± 0.01 | 0.936 ± 0.01 | 0.856 ± 0.01 |
| ODIN | 0.638 ± 0.01 | 0.983 ± 0.01 | 0.891 ± 0.02 | 0.774 ± 0.01 | 0.655 ± 0.01 | 0.900 ± 0.01 | **0.990 ± 0.01** | 0.792 ± 0.01 | 0.839 ± 0.01 | 0.567 ± 0.01 | 0.818 ± 0.01 | 0.677 ± 0.01 |
| ResAD | 0.919 ± 0.01 | 0.496 ± 0.01 | **0.907 ± 0.02** | 0.644 ± 0.01 | 0.882 ± 0.01 | 0.658 ± 0.01 | 0.811 ± 0.01 | 0.754 ± 0.01 | 0.839 ± 0.01 | **0.938 ± 0.01** | **0.960 ± 0.01** | **0.888 ± 0.01** |
| Our Model | **0.952 ± 0.01** | **0.984 ± 0.01** | 0.864 ± 0.01 | **0.853 ± 0.01** | 0.884 ± 0.01 | **0.980 ± 0.01** | 0.976 ± 0.01 | **0.930 ± 0.01** | 0.787 ± 0.01 | 0.900 ± 0.01 | 0.900 ± 0.01 | 0.881 ± 0.01 |

| Method | DDOS-HOIC (held-out) | | | | SLOWLORIS(meta-tune) | | | | INFILTRATION(held-out) | | | |
|---|---|---|---|---|---|---|---|---|---|---|---|---|
| | P | R | AUC | F1 | P | R | AUC | F1 | P | R | AUC | F1 |
| CORAL | 0.523 ± 0.01 | 0.843 ± 0.01 | 0.839 ± 0.01 | 0.645 ± 0.01 | **0.971 ± 0.01** | 0.733 ± 0.01 | 0.981 ± 0.01 | 0.835 ± 0.01 | 0.556 ± 0.01 | 0.569 ± 0.01 | 0.584 ± 0.01 | 0.562 ± 0.01 |
| MTAE | 0.497 ± 0.01 | 0.375 ± 0.01 | 0.505 ± 0.01 | 0.428 ± 0.01 | 0.783 ± 0.01 | 0.844 ± 0.01 | 0.841 ± 0.01 | 0.875 ± 0.01 | 0.688 ± 0.01 | 0.386 ± 0.01 | 0.546 ± 0.01 | 0.494 ± 0.01 |
| MTL-RED | 0.758 ± 0.01 | 0.534 ± 0.01 | 0.604 ± 0.01 | 0.627 ± 0.01 | 0.930 ± 0.01 | 0.773 ± 0.01 | 0.929 ± 0.01 | 0.844 ± 0.01 | 0.828 ± 0.01 | 0.584 ± 0.01 | 0.757 ± 0.01 | 0.684 ± 0.01 |
| ODIN | 0.787 ± 0.01 | 0.890 ± 0.01 | 0.890 ± 0.01 | 0.881 ± 0.01 | 0.690 ± 0.01 | 0.990 ± 0.01 | 0.823 ± 0.01 | 0.860 ± 0.01 | 0.571 ± 0.01 | 0.571 ± 0.01 | 0.684 ± 0.01 | 0.571 ± 0.01 |
| ResAD | **0.924 ± 0.01** | 0.483 ± 0.01 | 0.649 ± 0.01 | 0.634 ± 0.01 | 0.934 ± 0.01 | 0.716 ± 0.01 | 0.729 ± 0.01 | 0.810 ± 0.01 | **0.978 ± 0.01** | 0.044 ± 0.01 | 0.598 ± 0.01 | 0.084 ± 0.01 |
| Our Model | 0.862 ± 0.01 | **0.990 ± 0.01** | **0.926 ± 0.01** | **0.990 ± 0.01** | 0.787 ± 0.01 | **0.990 ± 0.01** | **0.990 ± 0.01** | **0.882 ± 0.01** | 0.718 ± 0.01 | **0.983 ± 0.01** | **0.830 ± 0.01** | **0.735 ± 0.01** |

| Method | DDoS (meta-tune) | | | | DoS (meta-tune) | | | | MIRAI (held-out) | | | |
|---|---|---|---|---|---|---|---|---|---|---|---|---|
| | P | R | AUC | F1 | P | R | AUC | F1 | P | R | AUC | F1 |
| CORAL | **0.986 ± 0.01** | 0.990 ± 0.01 | 0.980 ± 0.01 | **0.993 ± 0.01** | 0.978 ± 0.01 | 0.875 ± 0.01 | 0.961 ± 0.01 | 0.924 ± 0.01 | 0.819 ± 0.01 | **0.962 ± 0.01** | 0.863 ± 0.01 | 0.885 ± 0.01 |
| MTAE | 0.879 ± 0.01 | 0.990 ± 0.01 | 0.935 ± 0.01 | 0.964 ± 0.01 | 0.962 ± 0.01 | 0.990 ± 0.01 | **0.999 ± 0.01** | 0.982 ± 0.01 | 0.880 ± 0.01 | 0.653 ± 0.01 | 0.946 ± 0.01 | 0.750 ± 0.01 |
| MTL-RED | 0.956 ± 0.01 | 0.990 ± 0.01 | 0.990 ± 0.01 | 0.977 ± 0.01 | 0.974 ± 0.01 | **0.999 ± 0.01** | 0.990 ± 0.01 | 0.987 ± 0.01 | 0.913 ± 0.01 | 0.866 ± 0.01 | 0.859 ± 0.01 | 0.889 ± 0.01 |
| ODIN | 0.978 ± 0.01 | 0.833 ± 0.01 | 0.967 ± 0.01 | 0.900 ± 0.01 | 0.913 ± 0.01 | 0.866 ± 0.01 | 0.950 ± 0.01 | 0.889 ± 0.01 | 0.880 ± 0.01 | 0.653 ± 0.01 | 0.947 ± 0.01 | 0.780 ± 0.01 |
| ResAD | 0.948 ± 0.01 | 0.990 ± 0.01 | 0.990 ± 0.01 | 0.973 ± 0.01 | 0.930 ± 0.01 | 0.990 ± 0.01 | 0.993 ± 0.01 | 0.964 ± 0.01 | **0.934 ± 0.01** | 0.716 ± 0.01 | 0.729 ± 0.01 | 0.810 ± 0.01 |
| Our Model | 0.974 ± 0.01 | **0.990 ± 0.01** | **0.990 ± 0.01** | 0.987 ± 0.01 | **0.982 ± 0.01** | 0.990 ± 0.01 | 0.990 ± 0.01 | **0.990 ± 0.01** | 0.913 ± 0.01 | 0.866 ± 0.01 | **0.958 ± 0.01** | **0.889 ± 0.01** |

| Method | SPOOFING (held-out) | | | | RECONNAISSANCE(meta-tune) | | | | WEB (held-out) | | | |
|---|---|---|---|---|---|---|---|---|---|---|---|---|
| | P | R | AUC | F1 | P | R | AUC | F1 | P | R | AUC | F1 |
| CORAL | 0.519 ± 0.01 | 0.776 ± 0.01 | 0.553 ± 0.01 | 0.622 ± 0.01 | **0.989 ± 0.01** | 0.615 ± 0.01 | 0.912 ± 0.01 | 0.758 ± 0.01 | 0.741 ± 0.01 | 0.619 ± 0.01 | 0.666 ± 0.01 | 0.675 ± 0.01 |
| MTAE | 0.527 ± 0.01 | 0.788 ± 0.01 | 0.742 ± 0.01 | 0.632 ± 0.01 | 0.913 ± 0.01 | 0.603 ± 0.01 | 0.709 ± 0.01 | 0.726 ± 0.01 | 0.760 ± 0.01 | 0.628 ± 0.01 | 0.751 ± 0.01 | 0.688 ± 0.01 |
| MTL-RED | 0.913 ± 0.02 | 0.603 ± 0.01 | 0.709 ± 0.01 | 0.726 ± 0.01 | 0.890 ± 0.01 | **0.989 ± 0.01** | **0.988 ± 0.01** | **0.937 ± 0.01** | 0.824 ± 0.01 | 0.692 ± 0.01 | 0.883 ± 0.01 | 0.752 ± 0.01 |
| ODIN | 0.768 ± 0.01 | 0.414 ± 0.01 | 0.773 ± 0.01 | 0.538 ± 0.01 | 0.909 ± 0.01 | 0.822 ± 0.01 | 0.890 ± 0.01 | 0.864 ± 0.01 | 0.801 ± 0.01 | 0.603 ± 0.01 | 0.875 ± 0.01 | 0.723 ± 0.01 |
| ResAD | 0.632 ± 0.01 | **0.800 ± 0.01** | 0.762 ± 0.01 | 0.706 ± 0.01 | 0.909 ± 0.01 | 0.822 ± 0.01 | 0.890 ± 0.01 | 0.864 ± 0.01 | 0.882 ± 0.01 | 0.658 ± 0.01 | 0.811 ± 0.01 | 0.754 ± 0.01 |
| Our Model | **0.890 ± 0.02** | 0.716 ± 0.01 | **0.891 ± 0.02** | **0.794 ± 0.01** | 0.988 ± 0.01 | 0.825 ± 0.01 | 0.947 ± 0.01 | 0.899 ± 0.01 | **0.867 ± 0.01** | **0.860 ± 0.01** | **0.945 ± 0.02** | **0.863 ± 0.01** |

loop (meta-tuning), with some completely unseen(OOD) anomalies reserved for the test or inference phase. In the *inner loop*, we update the encoder and head using only ID (normal) data so the representation concentrates normal structure without being distracted by anomalies. In *outer loop*, we form a balanced ID-OOD query set using some disjoint set of anomalies (different from those held-out anomalies used at inference stage), and optimize a binary objective on the normal logit augmented with a hinge margin that explicitly enforces ID–OOD separation, and this process is repeated over multiple curriculums. Thus, the multi-directional flow of information arises from the various combinations of normal and anomaly sets used across the meta-learning episodes, which are then integrated to form the overall training curriculum with the final goal being generalizing to the completely unseen held-out anomaly classes. The temperature parameter is meta-tuned jointly with the classifier head to calibrate confidence margin between the ID(normal) and the OOD(anomaly) data. We follow a hard-OOD curriculum that prioritizes challenging families while cycling coverage across episodes. We select anomaly classes that have varied correlation patterns and are significantly different from the normal data samples. Selecting data from different correlation patterns provides sufficient coverage for better generalization. The train hyperparameters include the ID–OOD softmax calibration margin, which we fix at 0.05, the number of meta-learning episodes or tasks, the batch-size, the number of epochs, and the learning-rate for inner and outer training.

**Threshold Selection and Evaluation** After training, a single pooled decision threshold $\tau^\star \in [0, 1]$ is chosen on a validation pool by maximizing the F1 score:

$$\tau^\star \in \arg\max_{\tau \in [0,1]} \text{F1}\big(\mathbf{1}\{p(x; T, \theta_q^\star, \phi) \geq \tau\}, y \in I_q^c\big). \tag{8}$$

We then report the metrics (Precision, Recall, Accuracy, F1, Average Precision (AP), and AUC-ROC) on held-out and training classes at the end of each episode of the episodic multi-task meta-learning model using $\tau^\star$. We perform evaluation using a single global threshold (normal vs. all anomalies) for both the training OOD anomaly classes (used in meta-tuning) and the completely unseen held-out anomaly classes. Every dataset is balanced with an equal number of normal and anomaly samples. We select a single global threshold $\tau^*$ on a pooled ID/OOD validation set; the F1 score there is micro-averaged over all examples.

## 5 Datasets and Baselines

We perform exclusive experimentation on different cybersecurity intrusion detection datasets and in the healthcare domain.

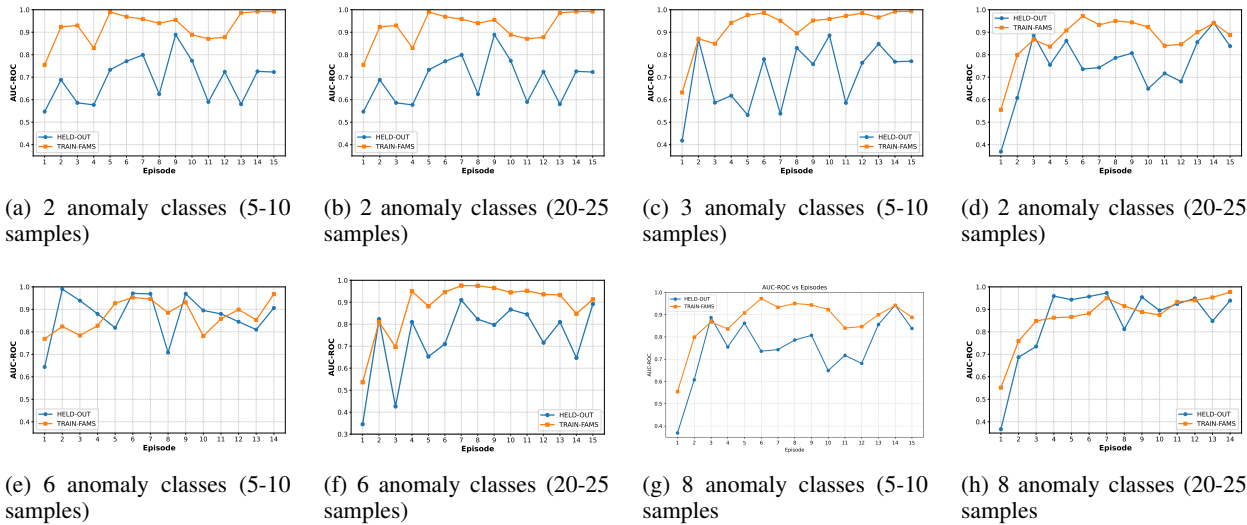

Figure 2: We plot the AUC-ROC against the number of episodes for both the held-out and the training (meta-tuning) anomaly classes. Across episodes, we gradually increase both the number of anomaly (OOD) classes included in the meta-tuning stage and the number of samples per class used for meta-tuning, allowing us to study how performance evolves as the meta-training curriculum becomes more diverse. Early stopping enables selection of a model that achieves strong performance on both held-out and training anomaly classes.

## 5.1 Dataset

**CSE-CIC-IDS2018 [Canadian-Institute, 2024]** This is a publicly available cybersecurity dataset that is made available by the Canadian Cybersecurity Institute (CIC). It consists of 7 major kinds of intrusion datasets namely Bruteforce, Web, Infiltration, Botnet, DoS, DDoS along with the Benign class.

**CICIoT 2023 [Canadian-Institute, 2024]** This is a state-of-the-art dataset for profiling, behavioral analysis, and vulnerability testing of different IoT devices with different protocols from the network traffic, consisting of 7 major attack classes - DoS, DDoS, Reconnaissance, Web, Mirai along with a Benign class for each attack categories.

**CICIoMT 2024 [Canadian-Institute, 2024]** This is a benchmark dataset to enable the development and evaluation of Internet of Medical Things (IoMT) security solutions. The attacks are categorized into five classes - DDoS, DoS, Reconnaissance, Spoofing, MQTT along with a Benign class for each attack categories.

**Arrythmia** This dataset is about atrial fibrillation (also called AFib or AF) which is a quivering or irregular heartbeat (arrhythmia) that can lead to blood clots, stroke, heart failure, and other heart-related complications. The dataset contains five classes/categories: N (Normal), S (Supraventricular ectopic beat), V (Ventricular ectopic beat), F (Fusion beat), and Q (Unknown beat).

Both CIC-IOT and CIC-IOMT datasets share a high degree of feature overlap and therefore we merge them based on their consistent feature sets to expand the number of task and class combinations available for multi-task meta-training.

## 5.2 Baselines

We consider the following recent models baselines covering zero-shot and few-shot domain generalization, multi-task learning and anomaly detection.

**Correlation Alignment for Deep Domain Adaptation (CORAL)** [Sun and Saenko, 2016] This work has been employed for supervised domain adaptation, aligning source and target covariances to enhance OOD generalization.

**Domain Generalization with Multi-task Autoencoders (MTAE)** [Ghifary et al., 2015] This encoder-decoder model optimizes reconstruction error across multiple domains in a supervised manner, jointly training sources and cross-domain data with label information in a two-stage process.

**Out-Of-Distribution Detection in Neural Networks (ODIN)** [Liang et al., 2017] The ODIN paper proposes two simple test-time tweaks—temperature scaling of the softmax and a small input perturbation that amplify the confidence gap between ID and OOD samples. For ODIN, we consider fixed temparature scaling value(T) and $\epsilon$ for the perturbation from a possible range of values.

**Improving Novel Anomaly Detection with Domain-Invariant Latent Representations (MTL-RED)** MTL-RED [Roy and Choi, 2025] propose a novel classification framework that leverages regularization techniques to guide the latent space toward retaining only the most relevant features for out-of-distribution (OOD) classification.

**A Simple Framework for Class Generalizable Anomaly Detection (Res-AD)** This paper [Yao et al., 2025] models residual features instead of raw features, which reduces interclass variation and keeps normal residuals consistent across classes. A detailed comparative study of our work and this recent work in the context of anomaly detection is presented in our supplementary material.

## 6 Ablation Study

Our ablation study mainly focuses on the components of our bilevel meta learner mainly calibrating the softmax ID-OOD margin with $\phi$ and occasionally $T$ (learned temperature vs fixed temperature). The margin penalty consistently sharpens ID-OOD separation and boosts AUC and AP. For our experiments, we also experiment with learning the temperature $T$ with a mild regularizer in order to improve calibration and ROC. Our configuration combines (across tasks and inner vs. outer), the margin penalty with $m = 0.5, \alpha = 0.1$, learned $T$ with and a frozen encoder, jointly maximizing AUC and improving per-class PR curves. While we explored learning the temperature parameter $T$, we observe significant improvement from meta-tuning the $\phi$ parameter to reshape the decision surface. We evaluate by rotating each dataset as held-out (OOD) while using only a small few-shot subset of anomalies from the remaining datasets (5-50 samples each class) for the upper-level meta-tuning; the encoder is first trained on normal traffic and then frozen for the inner stage. For testing, we construct class-balanced splits to ensure stable and comparable estimates across families and episodes.

Table 2: We report **AUC-ROC** (with std deviation) of the proposed and baseline methods on the Arrhythmia dataset. Each case uses equal amounts of normal and anomaly samples (balanced).

| Model | Anomaly (%) | TRAIN DOMAINS | | TRAIN DOMAINS | | OOD DOMAINS |
|---|---|---|---|---|---|---|
| | | VEB | BENIGN | SVEB | Q | F |
| MTAE | 50% | 0.99 (0.12) | 0.63 (0.16) | 0.61 (0.15) | 0.90 (0.12) | 0.60 (0.15) |
| CORAL | 50% | 0.98 (0.10) | 0.58 (0.10) | 0.68 (0.18) | 0.96 (0.13) | 0.73 (0.18) |
| MTLS-RED | 50% | **0.99 (0.05)** | 0.78 (0.11) | 0.84 (0.18) | 0.96 (0.23) | **0.83 (0.13)** |
| Our Model | 50% | **0.99 (0.05)** | **0.89 (0.11)** | **0.99 (0.11)** | 0.99 (0.10) | **0.89 (0.13)** |

Overall, Tables 1–2 and the episodic analysis in the ablation with our multidirectional meta-learning yields the largest gains on hard, distribution-shifted OOD families (e.g., BOTNET, GOLDENEYE, INFILTRATION, SVEB,Q,F(OOD) in Arrhythmia), where F1 and AUC (approximately 15–30%) improve substantially over CORAL, MTAE, MTL-RED, the ODIN baseline, and RESAD, while yielding only marginal improvements on easier families such as DoS/DDoS. Compared to the recent RESAD framework, we find that our multidirectional meta-learning model consistently achieves higher F1 scores on the hardest attack families.

In Figure 2, we plot AUC-ROC as a function of the number of episodes for both held-out and training (meta-tuning) anomaly classes. As training progresses, we

---
[0]GitHub Code repository

---

Table 3: Ablation comparing meta-tuning of classifier head $\phi$ and $T$ under different few-shot OOD budgets per class (meta-tuning).

| Few-shot Case | Meta-tune | Held-out F1 | Held-out AUC | Meta OOD F1 | Meta OOD AUC | Held-out Margin | Meta Margin |
|---|---|---|---|---|---|---|---|
| 5 samples | $T, \phi$ | 0.805 | 0.821 | 0.830 | 0.965 | 0.073052 | 0.397226 |
| 20 samples | $T, \phi$ | 0.745 | 0.797 | 0.920 | 0.965 | 0.125505 | 0.203727 |
| 50 samples | $T, \phi$ | 0.813 | 0.846 | 0.972 | 0.993 | 0.000275 | 0.900369 |
| 5 samples | $T, \phi$ | 0.890 | 0.899 | 0.786 | 0.864 | 0.623420 | 0.431433 |
| 20 samples | $T, \phi$ | 0.793 | 0.774 | 0.748 | 0.775 | 0.001882 | 0.000164 |
| 50 samples | $T, \phi$ | 0.906 | 0.939 | 0.924 | 0.977 | 0.002842 | 0.323710 |

gradually increase both the number of anomaly (OOD) classes introduced during meta-tuning and the number of samples per class, enabling analysis of how performance evolves as the meta-training curriculum becomes increasingly diverse. We observe that as we gradually introduce more diversity in the outer-level meta-tuning process, the model generalization to completely unseen anomaly classes improves. Early stopping allows selecting a best model that performs well on both held-out and training anomaly classes.

In Table 3, we fix the inner-trained encoder/decoder and run the outer meta-loss with 5/20/50 few-shot OOD samples per meta-OOD class, comparing meta-tuning the classifier head $\phi$ and occasionally $T$ on the held-out and meta-tune anomaly classes using a *global threshold* calibrated individually on them. Tuning ($T$) does not yield any significant gain, whereas, tuning $\phi$ contributes significantly to calibrating the decision surface. With 50-shot and meta-tuning $\phi$, we obtain the strongest overall improvements, both in accuracy and the softmax confidence margin. Smaller thresholds can be numerically fragile or essentially an over-confident model, for e.g, small shifts in calibration or noise can flip the ID-OOD margin.

## 7 Conclusion

We presented a bilevel training framework for zero-shot anomaly detection that disentangles an inner supervised objective for representation learning with an outer softmax calibration objective. The outer stage enforces an adaptive margin on calibrated logits. Across episodes using different combinations of domain splits, we aim to achieve domain generalization to completely unseen (OOD) anomalies while maintaining equally good segregation of the known (train) anomaly classes. Empirically, most OOD anomaly families benefit from the two-stage disentanglement scheme, which yields higher AUC and demonstrates the usefulness of the margin-based calibration (meta-tuning) using a small number of anomaly classes. We conduct a wide range of experiments on a number of standard cybersecurity and healthcare datasets to showcase the improvements.

# A  Bi-level Meta-Optimization for Out-Of-Distribution (OOD) Generalization

We consider a representation learner composed of an encoder–decoder pair and a classifier head. Let $f_{\theta_{\text{enc}}} : \mathbb{R}^D \to \mathbb{R}^m$ be the encoder, $g_{\theta_{\text{dec}}} : \mathbb{R}^m \to \mathbb{R}^D$ the decoder, and $h_\phi : \mathbb{R}^m \to \mathbb{R}^2$ the classifier producing logits. We collect the encoder–decoder parameters as $\theta = (\theta_{\text{enc}}, \theta_{\text{dec}})$ and the classifier parameters as $\phi$. Given an input $x \in \mathbb{R}^D$, the forward map is

$$z = f_{\theta_{\text{enc}}}(x), \qquad \hat{x} = g_{\theta_{\text{dec}}}(z), \qquad \ell = h_\phi(z) \in \mathbb{R}^2,$$
$$p = \left(\tfrac{1}{T}\,\ell\right).$$

where $T > 0$ is a temperature (learned in the outer loop; see below).

## A.1  Inner Objective (Lower Level)

**Normals-only one-class training** The inner objective now *uses only ID (normal) data* to shape the representation and head. We maximize confidence on normals with a one-class binary cross-entropy, while keeping the variance-normalized reconstruction term to a very minimum value.

$$(\theta, \phi;^{\text{ID}}) =_{x \sim^{\text{ID}}} \left[\left(p_0(x; T{=}1),\, 1\right)\right] \quad (9)$$

where $p_0(x; T{=}1) = (\ell(x))_0$, and $\lambda_{\text{rec}} \geq 0$

Gradients w.r.t. $\phi$ arise from the one-class BCE, while gradients w.r.t. $\theta$ arise from the same BCE (via the encoder) and optionally from reconstruction:

$$\frac{\partial}{\partial \phi} = \left[\frac{\partial}{\partial \ell}\,\frac{\partial h_\phi}{\partial \phi}\right], \qquad \frac{\partial}{\partial \theta} = \left[\frac{\partial}{\partial \ell}\,\frac{\partial h_\phi}{\partial z}\,\frac{\partial f_{\theta_{\text{enc}}}}{\partial \theta}\right]$$

Denoted by
$$\theta^\star(\phi) \in\ _\theta\ (\theta, \phi;^{\text{ID}}) \quad (10)$$
the (possibly set-valued) inner solution for a fixed $\phi$.

## A.2  Outer Objective (Upper Level): BCE with Meta OOD (Anomalies)

**Anomalies only in the outer loop** The meta-objective contrasts ID (normals) against meta OOD (anomalies) on an *outer* query set $=^{\text{ID}} \cup^{\text{OOD}}$, disjoint in anomaly classes from $^{\text{ID}}$. We replace the previous softplus-margin with a calibrated BCE that encourages high confidence on ID and low confidence on OOD; the temperature $T$ and decision boundary $\phi$ are meta-tuned:

$$(T, \phi; \theta^\star) =_{x \sim^{\text{ID}}} \left[\left(s(x; \theta^\star, \phi, T),\, 1\right)\right]$$
$$+_{x \sim^{\text{meta-OOD}}} \left[\left(s(x; \theta^\star, \phi, T),\, 0\right)\right] \quad (11)$$

where $\theta^\star = \theta^\star(\phi)$ as in Eq 1. This keeps the calibration mechanism in the meta level, while the inner level models normality.

Although both are "parameters," they *enter different parts of the computation*: $\theta$ shapes the representation $z = f_{\theta_{\text{enc}}}(x)$ and reconstruction $g_{\theta_{\text{dec}}}$, hence the *geometry* of latent space; $\phi$ tilts the *decision surface* $h_\phi(z)$ over that geometry. This separation yields different gradient pathways in the bi-level formulation.

## A.3  Hypergradients and First-Order Approximation

For a true bi-level update of $\phi$, differentiating through the inner optimum $\theta^\star(\phi)$ yields

$$\frac{d}{d\phi} = \underbrace{\frac{\partial}{\partial \phi}}_{\text{direct (logits) path}} + \underbrace{\frac{\partial}{\partial \theta^\star}\,\frac{d\theta^\star}{d\phi}}_{\text{implicit (hypergradient) path}}. \quad (12)$$

By the implicit function theorem,

$$\frac{d\theta^\star}{d\phi} = -\left[\nabla_\theta^2(\theta^\star, \phi)\right]^{-1} \nabla_\phi \nabla_\theta(\theta^\star, \phi). \quad (13)$$

Computing and inverting the inner Hessian is costly and can be numerically brittle. In practice, we employ a *first-order* approximation that *freezes* $\theta$ at $\theta^\star$ during the outer update and optimizes only $(T, \phi)$. This removes the implicit term and keeps only the direct, fully differentiable path through the logits:

$$\frac{d}{d\phi} \approx \frac{\partial}{\partial \phi} \quad \text{with } \theta = \theta^\star \text{ treated as constant.} \quad (14)$$

Because both the score $s$ and the perturbation direction $d$ depend on the logits $h_\phi$, the outer gradient still meaningfully adapts $\phi, T$ to enlarge the ID–OOD gap, while preserving the inner-learned representation.

**Role of Representation and Classifier in Bi-level Meta-Tuning**

Although both $\theta$ and $\phi$ are parameters of the model, they enter different parts of the computation and therefore play fundamentally different roles in separating ID and OOD data. The representation parameters $\theta$ determine the mapping $z = f_\theta(x)$ (and, in the autoencoding variant, the reconstruction $g_\theta(z)$), and thus control the *geometry* of the latent space $\mathcal{Z}$. Training the inner objective on many normal families shapes $\mathcal{Z}$ into a tight "normal manifold" $f_\theta(\text{ID})$, where samples from all seen normal classes cluster in a high-density region and OOD samples naturally fall into low-density, off-manifold regions. In contrast, the classifier head $\phi$ only acts on latent variables through the decision function $h_\phi(z)$ and its associated calibrated score $s(x; \theta, \phi, T)$ (e.g., an anomaly score derived from the max-softmax of $h_\phi(f_\theta(x))$). In the bi-level formulation, the inner loop primarily updates $\theta$ (and an initial $\phi^{(0)}$) to obtain a geometry in which normal data are well organized; the outer loop then freezes $\theta$ and meta-tunes $\phi$ using a small number of anomaly examples so as to tilt and rescale the decision surface $h_\phi$ over this fixed geometry. Intuitively, gradients flowing through $\theta$ carve out the normal manifold ID, while gradients flowing through $\phi$ sharpen the boundary between ID and OOD (anomaly): points near the manifold are pushed toward uniformly high normal confidence, whereas off-manifold points are assigned uniformly low confidence. This separation of geometric (via $\theta$) and decision-surface (via $\phi$) pathways is precisely what enables the bi-level optimization to enlarge the ID–OOD margin using only light, few-shot anomaly supervision in the outer loop.

### A.4 Gradients w.r.t. Meta-Parameters

Let $y \in \{0,1\}$ denote the ID/OOD label on ($y=1$ for ID), and $s(x)$ as above. For any meta-parameter $\psi \in \{T, \phi\}$,

$$\frac{\partial \mathcal{L}}{\partial \psi} = \mathbb{E}_{(x,y)\sim}\left[(s(x) - y)\frac{\partial s(x)}{\partial \psi}\right] + \frac{\partial \mathcal{R}}{\partial \psi}, \quad (15)$$

where $\frac{\partial s}{\partial \psi}$ is obtained by automatic differentiation through the chain $x \to d(x; \cdot) \to f_{\theta_\text{enc}} \to h_\phi \to (\cdot/T) \to \max$. As before, we use a smooth surrogate for the $\max$ (e.g., $\log\sum_k e^{\cdot}$ or a straight-through top-1) and a single-step (one backward pass per batch) to stabilize gradients.

### A.5 Why Bi-level Helps OOD

The inner objective Eq 3 learns a robust latent geometry of *normal* patterns (without using anomalies), optionally regularized by reconstruction. The outer objective Eq 4 then *calibrates* the logits and meta-parameters ($T, \phi$ to *maximize* separability on *disjoint* (unseen) anomaly families via BCE. This two-stage pressure (shape normality; then widen ID(normal)–OOD(anomaly) margins under input perturbations) improves generalization to unseen/OOD anomalies by aligning the classifier confidence landscape and its input gradients with the downstream detection criterion.

### A.6 Effect of Meta-Tuning the Decision Boundary with Meta-OOD

We view the proposed training scheme as a bi-level optimization problem with a representation map $f_\theta : \mathcal{X} \to \mathcal{Z}$ and a classifier head $h_\phi : \mathcal{Z} \to \mathbb{R}^K$. Let $\mathcal{D}_\text{ID}$ denote the union of all *normal* families used in the inner loop and $\mathcal{D}_\text{meta}$ denote the few-shot meta-episodes composed of pairs $(x, y) \in \{0, 1\}$ where $y = 1$ encodes ID/normal and $y = 0$ encodes OOD/anomaly. The inner loop optimizes

$$(\theta^\star, \phi^{(0)}) \in \arg\min_{\theta, \phi} \mathbb{E}_{(x,y)\sim\mathcal{D}_\text{ID}} \ell_\text{CE}(h_\phi(f_\theta(x)), y),$$

so that $f_{\theta^\star}$ induces a compact "normal manifold" $\mathcal{M}_\text{ID} = f_{\theta^\star}(\text{supp}(\mathcal{D}_\text{ID})) \subset \mathcal{Z}$, and $h_{\phi^{(0)}}$ already separates this manifold from the complement of $\mathcal{Z}$ in a standard supervised sense. Given $\theta^\star$, the outer loop freezes the geometry and only meta-tunes $\phi$ (and, optionally, a scalar temperature $T > 0$) using a calibrated anomaly score

$$s(x; \phi, T) = 1 - p_\text{ID}(x; \phi, T) = 1 - \sigma\left(\frac{1}{T}h_\phi(f_{\theta^\star}(x))\right)_{k_\text{ID}},$$

where $\sigma$ is the softmax and $k_\text{ID}$ indexes the normal class. The meta-objective is

$$(\phi^\star, T^\star) \in \arg\min_{\phi, T} \mathbb{E}_{(x,y)\sim\mathcal{D}_\text{meta}} \ell_\text{meta}(s(x; \phi, T), y),$$

with $\ell_\text{meta}$ chosen to enlarge the margin $\mathbb{E}[s(x; \phi, T) \mid y = 0] - \mathbb{E}[s(x; \phi, T) \mid y = 1]$. Crucially, the gradients of this outer objective flow only through $\phi$ and $T$; $\theta^\star$ remains fixed and continues to encode the shared normal manifold. In practice we observe that the dominant effect comes from the update $\phi^{(0)} \mapsto \phi^\star$, which rotates and rescales the decision surface $h_\phi \circ f_{\theta^\star}$ over the fixed geometry (shared normal) so that points near the normal manifold attain uniformly low anomaly scores $s(x; \phi^\star, T^\star)$ while off-manifold points attain uniformly high scores. The primary contribution of the outer loop is to sharpen the global decision boundary via $\phi$ and the temperature $T$ playing a secondary calibration role

**Temperature calibration and scoring.** We always compute $g(x) = \ell_0(x)$ using *calibrated* logits $\ell(x) = h_\phi(f_{\theta_\text{enc}}(x))/T$ both when forming the direction Eq (1) and when evaluating the perturbed input for the loss. The final anomaly scores use a sigmoid over the calibrated normal logit, $p_\text{norm}(x) = \sigma(\ell_0(x))$, ensuring consistency between training and evaluation.

We (i) solve the inner problem for $(\theta, \phi)$ on episodic ID-only mixtures, (ii) freeze $\theta = \theta^\star$ and optimize $(T, \phi)$ on disjoint families via Eq (6), and (iii) calibrate a global threshold on a balanced validation mix for reporting (Precision, Recall, Accuracy, F1, AP/AUC-PR, AUC-ROC). This first-order scheme is stable, fully differentiable in $(T, \phi)$, and empirically widens the score gap between ID and OOD.

Figure 3 illustrates the UMAP projections of validation samples in the learned latent space. Panels (a), (c), and (e) correspond to anomaly classes used during meta-tuning, while panels (b), (d), and (f) show the latent representations of the held-out anomaly classes. For all experiments, the latent space dimensionality is fixed to 128.

### A.7 Neural Network Architecture

Our proposed bi-level meta-learning framework uses a feedforward encoder–decoder–classifier architecture –

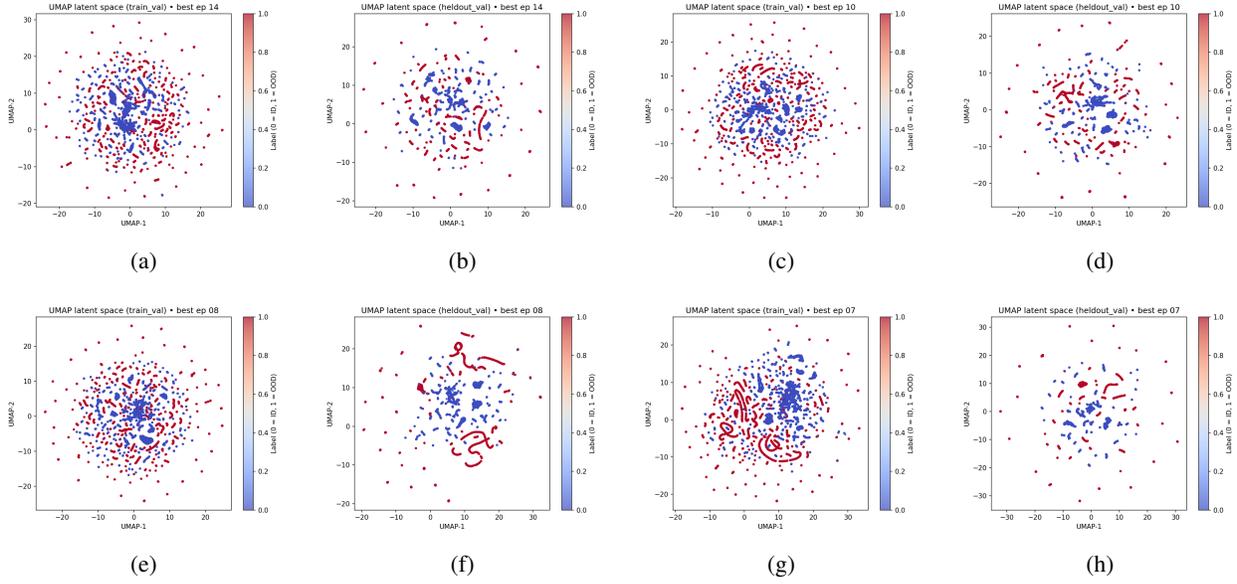

Figure 3

the encoder $f_\theta$ is a stack of residual fully connected blocks with LayerNorm and GELU activations that maps inputs into a bottleneck latent space, the decoder reconstructs the original features from this bottleneck, and a shallow classifier head $h_\phi$ (LayerNorm + linear layer) operates on the latent representation to produce normal/anomaly logits. Crucially, the meta-objective only interacts with the model through $f_\theta$ and $h_\phi$, and does not depend on any specific architectural details of these components. As a result, the multi-directional meta-learning procedure is agnostic to the neural backbone and can be instantiated with alternative dense architectures, such as deeper residual MLPs, ResNets, or vision transformers when working with image data, simply by substituting the encoder and classifier modules while keeping the bi-level optimization unchanged.